\documentclass{article}
\usepackage{tabularx}
\usepackage{graphicx}
\usepackage{wrapfig}
\usepackage[linesnumbered,ruled,vlined]{algorithm2e}
\usepackage[preprint]{corl_2024} 
\usepackage[utf8]{inputenc} 
\usepackage[T1]{fontenc}    
\usepackage{hyperref}       
\usepackage{url}            
\usepackage{booktabs}       
\usepackage{amsfonts}       
\usepackage{nicefrac}       
\usepackage{microtype}      
\usepackage{xcolor}         

\hypersetup{
    colorlinks=true,
    linkcolor=orange,
    filecolor=magenta,      
    urlcolor=orange,
    citecolor=orange,
}

\usepackage{titlesec}
\usepackage{xspace}
\usepackage{mathtools}
\usepackage{bbm}
\usepackage{enumitem}
\usepackage{caption}
\usepackage{subcaption}
\usepackage{multirow}
\usepackage{pdflscape}
\usepackage{xcolor}
\usepackage[]{minted}
\usepackage[toc,page,header]{appendix}
\usepackage{minitoc}

\setcitestyle{numbers,square,comma}
\usepackage{amsthm}
\usepackage{amsmath}

\usepackage[capitalise, nameinlink]{cleveref}

\newcommand{\ACRO}{\textsc{gamma}\xspace}

\title{Intent at a Glance: Gaze-Guided Robotic Manipulation via Foundation Models}

\makeatletter
\def\thanks#1{\protected@xdef\@thanks{\@thanks
        \protect\footnotetext{#1}}}
\makeatother

\author{
Tracey Yee Hsin Tay$^{\heartsuit}$, 
Xu Yan$^{*}$,
Jonathan Ouyang$^{*}$, Daniel Wu, William Jiang, \\[5pt]
\textbf{Jonathan Kao, Yuchen Cui$^{\dagger}$\thanks{{
\noindent $^{\heartsuit}$ Work done as an exchange student at UCLA.
$^{*}$ Equal contribution. \\
$^{\dagger}$ Corresponding Author. Email: {\texttt{yuchencui@cs.ucla.edu}}
}}
}
\\[5pt]
University of California, Los Angeles
\vspace{-0.5cm}
}

\begin{document}
\maketitle

\begin{abstract}
    Designing intuitive interfaces for robotic control remains a central challenge in enabling effective human-robot interaction, particularly in assistive care settings. Eye gaze offers a fast, non-intrusive, and intent-rich input modality, making it an attractive channel for conveying user goals. In this work, we present \ACRO (Gaze Assisted Manipulation for Modular Autonomy), a system that leverages ego-centric gaze tracking and a vision-language model to infer user intent and autonomously execute robotic manipulation tasks. By contextualizing gaze fixations within the scene, the system maps visual attention to high-level semantic understanding, enabling skill selection and parameterization without task-specific training. We evaluate \ACRO on a range of table-top manipulation tasks and compare it against baseline gaze-based control without reasoning. Results demonstrate that \ACRO provides robust, intuitive, and generalizable control, highlighting the potential of combining foundation models and gaze for natural and scalable robot autonomy. 
    Project website: \url{https://gamma0.vercel.app/}
\end{abstract}

\section{Introduction}

As robots become more integrated into human environments, enabling seamless collaboration between humans and robots is a growing priority across domains such as assistive care, household assistance, and human-in-the-loop industrial settings. For individuals with physical disabilities, robotic systems hold the promise of restoring agency and independence in daily tasks. However, realizing this potential requires intuitive and accessible interfaces—particularly for users who may have limited mobility or dexterity. Traditional control modalities, such as joysticks or touchscreens, often impose cognitive or physical burdens that are prohibitive for many users.

This challenge has spurred growing interest in multimodal interfaces that allow more natural and low-effort interaction~\cite{aronson2022gaze,lin2023giraf}. Among them, eye gaze has emerged as a particularly compelling modality—it is fast, non-intrusive, and deeply intuitive, as where we look often precedes and reflects our intentions~\cite{aronson2022gaze,chang2023specifying}. For users with motor impairments, gaze can serve as a powerful alternative to physical input, enabling control and communication without requiring hand or body movement~\cite{ward2000dasher,feng2021hgaze}. Gaze provides not only a predictive signal of upcoming actions but also a fine-grained indication of user goals; for example, looking at the body of a bottle versus its cap can imply different intended manipulations. These qualities make gaze a uniquely valuable channel for conveying intent in human-robot interaction, especially in assistive contexts.

\begin{figure}
    \includegraphics[width=\linewidth]{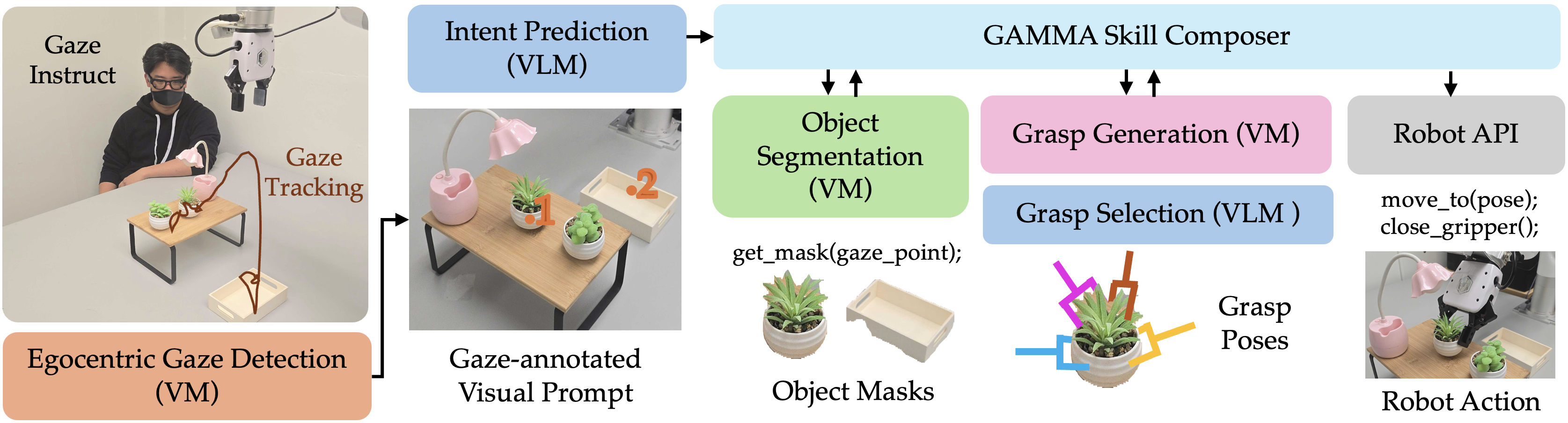}
    \caption{\textbf{Overview of \ACRO.} A user wearing smart glasses uses their gaze to specify a manipulation task. In this example, the user wants the robot to pick up a plant under the lamp and place it in a tray. \ACRO transforms the gaze fixations into robot's view and prompts the VLM to predict user intent. Given the predicted user intent, \ACRO calls corresponding functions for perception, planning, and execution. \ACRO prompts a VLM to select a proper grasping pose that takes the task context into consideration (e.g. not colliding with the lamp). }
    \label{fig:overview}
    \vspace{-0.3cm}
\end{figure}

In this work, we explore the use of gaze fixations as a window into user intent for autonomous robotic manipulation. Our key insight is that human gaze, when interpreted through the lens of powerful pretrained foundation models~\cite{bommasani2021opportunities}, can serve as an effective bridge between low-level sensor data and high-level task understanding. Foundation models trained on large-scale multimodal data bring strong generalization capabilities and semantic grounding, making them ideal candidates for interpreting gaze and planning for motion in visually complex and dynamic environments.

We propose an integrated assistive manipulation system named \ACRO (Gaze Assisted Manipulation for Modular Autonomy). An overview of \ACRO is illustrated in \cref{fig:overview}. \ACRO combines real-time gaze tracking from lightweight wearable devices with advanced pretrained vision models (VMs) and vision-language models (VLMs) to accurately interpret user intent and facilitate targeted manipulation tasks. By analyzing user gaze patterns and identifying fixation points within the observed environment, \ACRO leverages a diverse set of foundation models to understand and contextualize these points of interest. Subsequently, the robot autonomously determines the appropriate robotic skills and parameters to execute tasks aligned with the user's inferred intentions.
This removes the need for handcrafted intent prediction models or task-specific training, allowing for flexible, scalable deployment in real-world settings.

We evaluated \ACRO on a set of table-top manipulation tasks, including watering a plant, selecting objects from clutter, and making coffee. Our experiments compared \ACRO to a panel-based gaze control baseline. Results demonstrate that \ACRO enables intuitive, low-effort interaction and supports zero-shot manipulation through gaze and foundation model reasoning. \ACRO takes less than half the amount of time the user needs to spend with the baseline method.
Despite \ACRO's advantages in efficiency and reduced user effort, participants expressed a preference for the baseline method, citing its greater sense of control over the robot's actions.
These findings highlight an important consideration in the design of assistive systems: \textit{users value not only automation and ease of use, but also agency and the ability to influence robot behavior more directly}. This work contributes to the growing field of human-centered robotics by demonstrating how VLMs and gaze input can be combined to create intelligent, natural, and empowering assistive robotic systems—while also surfacing compelling challenges in balancing autonomy with user control.
\section{Related Work}

Gaze is a key non-verbal cue for attention and has been widely studied in various settings of human-robot interaction and robot learning~\cite{admoni2017social,saran2018human,shi2021gazeemd,belardinelli2024gaze}. We focus on discussing gaze-based methods for intent recognition and manipulation.
In parallel, a growing body of recent work is exploring leveraging foundation models for robotics~\cite{firoozi2023foundation,kawaharazuka2024real}. Here, we focus on those closely related to designing effective and natural human-robot interaction interfaces.

\noindent \textbf{Gaze-Based Interaction and Assistive Robotics.}
A substantial body of work has investigated gaze as an input modality in assistive multimodal teleoperation, demonstrating its effectiveness in complementing—and in some cases surpassing—traditional controls for goal recognition~\cite{admoni2016predicting,shafti2019gaze,aronson2021inferring,aronson2022gaze,fuchs2021gaze,prada2023gaze,chang2023specifying}.
While previous approaches have leveraged gaze for enhancing robot intention inference, they often required additional input devices such as joysticks for direct control. 
This limits the broader applicability of gaze-based control, especially in assistive settings for individuals with motor impairments, where gaze has proven vital for enabling interaction with computer systems~\cite{ward2000dasher,feng2021hgaze}.
\citet{wang2018free} demonstrated that free-view gaze control offers a more intuitive user experience, though their system was constrained to planar pick-and-place tasks.
Extending gaze control to full 6 DoF manipulation remains challenging for two key reasons. First, gaze provides only point estimates, making it difficult to specify orientation; achieving full 3D control often necessitates on-screen rotation interfaces or low-level robot primitives guided by intent prediction~\cite{wang2020toward,wang2023gaze}. 
Second, gaze-based intent inference is inherently ambiguous and typically requires additional contextual reasoning to accurately interpret user goals.
In contrast, our system, \ACRO, harnesses the reasoning capabilities of modern vision-language models (VLMs) to enable direct zero-shot robot manipulation using gaze alone. It eliminates the need for physical controllers or hard-coded, object-specific policies. 
\ACRO combines off-the-shelf computer vision models for object segmentation and grasp prediction with VLM-based reasoning to support zero-shot, gaze-driven manipulation.

\noindent \textbf{Foundation Models for Human-Robot Interaction.}
Recent advances in foundation models, especially large language models (LLMs), have significantly impacted human-robot interaction (HRI), enabling improved interpretation of human gestures \cite{lin2023giraf}, facilitating natural language-based corrections \cite{zha2024distilling}, and supporting adaptive skill composition through interactive learning \cite{grannenvocal}. 
Recent work of \citet{nasiriany2024pivot} pioneered the idea of using iterative visual prompts for zero-shot planning of robotics tasks. \ACRO is inspired by this idea and applies visual prompting to both intent recognition and low-level grasp selection.
Concurrent research also explores combining gaze with language models for assistive manipulation but has predominantly focused on high-level task reasoning \cite{lai2025fam,zhang2025mindeye}.
Distinctively, our approach harnesses the comprehensive capabilities of vision-language models for both high-level intent inference and detailed low-level motion planning tasks, including context-aware grasp selection. 
This integration enables \ACRO to provide a more flexible, scalable, and robust assistive robotic system suitable for diverse real-world applications.

\section{\ACRO: Gaze Assisted Manipulation for Modular Autonomy}
\label{sec:method}

\begin{figure}[b]
    \centering
    \includegraphics[width=\linewidth]{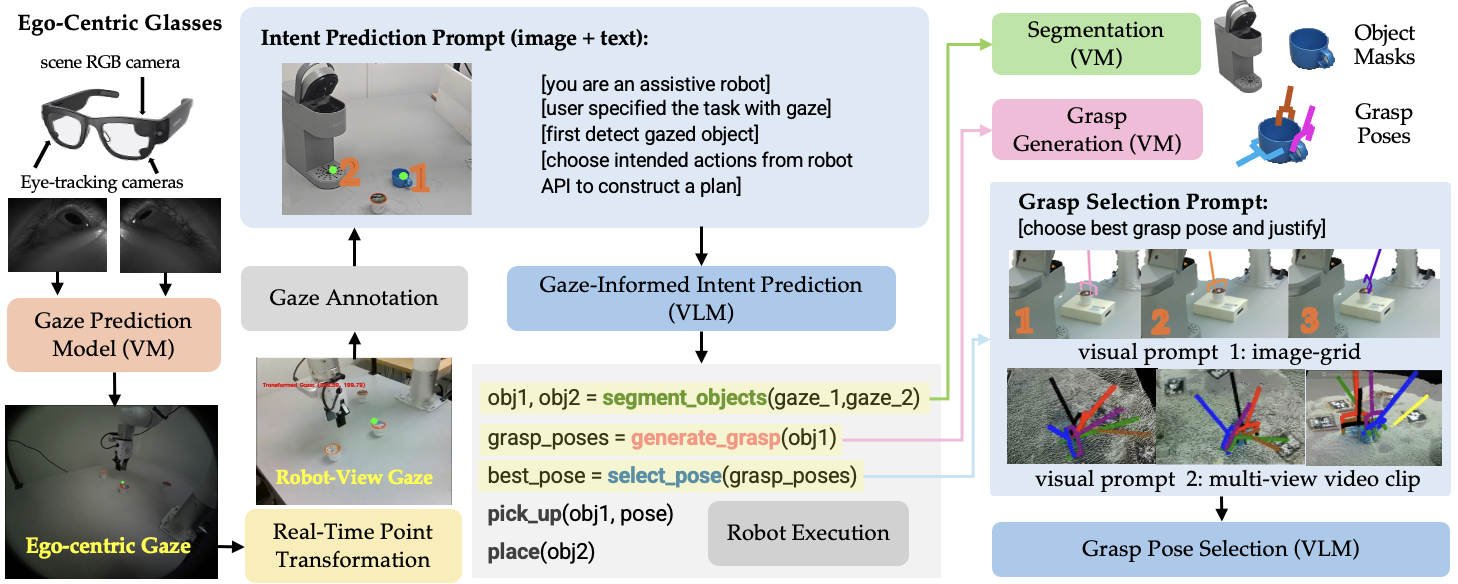}
    \caption{\textbf{Functional Modules of \ACRO.} \ACRO consists of various sensing \& perception modules that leverages pretrained vision models, and VLM-based reasoning modules. }
    \label{fig:system}
    \vspace{-0.25cm}
\end{figure}

We propose \ACRO as a general framework for controlling a robotic arm to perform manipulation tasks using egocentric eye gaze. 
We introduce key functional modules required for zero-shot manipulation and integrate off-the-shelf vision and vision-language foundation models for both perception and reasoning to enable modular, zero-shot autonomy.  \cref{fig:system} illustrates the key functional modules of \ACRO. We detail the design of each module in the following sections.

\subsection{The \ACRO Framework}

\ACRO follows a general structure for interpreting gaze-driven manipulation tasks. \ACRO assumes access to a set of parameterized robot APIs that contain both high-level skills such as \texttt{open(door)}, and low-level commands such as \texttt{go\_to(pose)}. 
Given a predicted intent $I_g$ from a vision-language model (VLM), the skill composer generates a behavior program $\mathcal{B}(I_g)$. We assume that each gaze instruction $I_g$ consists of gaze fixations on a sequence of target objects and can be compiled into a plan $\mathcal{P}$: a sequence of atomic skills $s(o, a)$, where each skill applies a parameterized motion primitive $a$ to a target object $o$.
Crucially, the order of gaze targets does not necessarily match the execution order required by the task. For instance, if a user gazes at a glass of milk and then at a microwave to indicate the goal of placing the milk inside the microwave, the robot must first open the microwave—corresponding to the second gaze target. As such, robust task planning through VLM-based reasoning is essential to correctly infer and execute the intended sequence of actions. To allow effective behavior planning, we design parameterized APIs for VLM planners to interface with perception and execution modules. Intent prediction and grasp selection are essentially cascaded multiple choice questions.
For perception modules, we leverage a 2D vision model for object segmentation and a 3D-based grasp prediction with pointcloud as input.

\begin{figure}
    \centering
    \includegraphics[width=\linewidth]{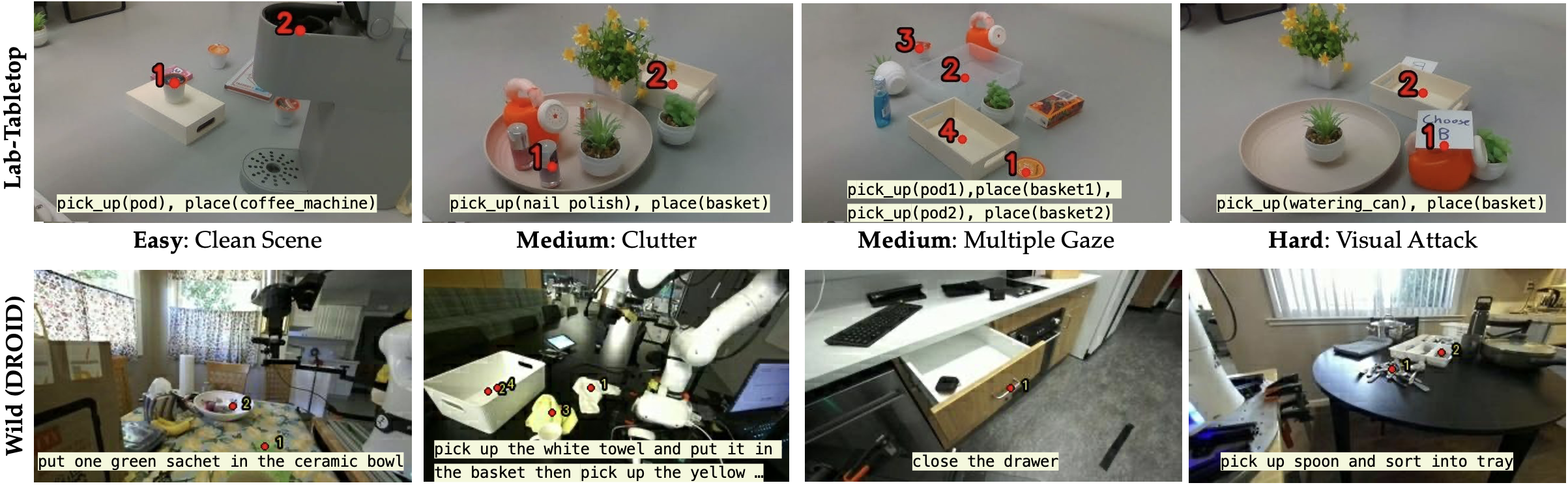}
    \caption{\textbf{Gaze-based Intent Reasoning Tasks.} (Top) We designed 30 tabletop manipulation scenarios in lab for intent reasoning with diverse difficulty levels. Easy scenes are relatively clean, medium difficulty-level scenes are cluttered or contain longer sequences of gaze points. The hard cases involves visual attacks. (Bottom) We also randomly sampled 45 scenes from the DROID~\cite{khazatsky2024droid} dataset containing in-the-wild manipulation tasks and annotated the scenes for intent reasoning.}
    \label{fig:intent-reasoning}
    \vspace{-0.25cm}
\end{figure}

\subsection{Sensing and Perception Modules}

\ACRO consists of sensing modules for capturing user gaze and perception modules for extracting scene information to support downstream reasoning and execution. To enable these capabilities, we integrate a suite of pretrained vision models tailored to specific sub-tasks, including gaze estimation, object segmentation, and grasp generation. This modular design allows \ACRO to flexibly interpret user gaze and transform it into actionable robot commands without requiring task-specific data.

\noindent\textbf{Gaze Detection and Transformation.}
Recent advances in wearable hardware enable precise 3D eye-tracking using commercially available devices. In \ACRO, we leverage the Aria glasses from Meta's Project Aria \cite{projectaria}. Project Aria's pretrained gaze estimation model predicts gaze points within the user's egocentric scene camera view using inputs from integrated eye-tracking cameras (\cref{fig:system}, left).
While instantaneous gaze measurements can be noisy and unreliable for inferring human intent, \ACRO aggregates gaze data over time to produce a more stable and robust intent signal. However, since the egocentric camera is constantly moving, the captured images change rapidly. To accurately ground human gaze in the relatively static environment, we transform gaze points from the egocentric frame into the robot's camera viewpoint using feature-based real-time camera pose estimation.

In practice, we employ ArUco markers as reference features, providing an efficient solution for gaze mapping between the user and the robot's shared workspace. However, transforming a 2D gaze point from the egocentric view into a third-person perspective is inherently ill-posed, as it corresponds to a ray rather than a single point.
Project Aria's 3D gaze model is also not sufficiently accurate for our application. Accurate gaze estimation requires computing the intersection of the gaze ray with the 3D point cloud from the robot’s viewpoint—a process that is computationally expensive to perform on every frame. As a practical workaround, we ask the user to remain close to the robot’s camera viewpoint to minimize transformation error and improve localization accuracy.


\noindent\textbf{Object Segmentation and Identification.}
Rather than relying on explicit object detection or classification, \ACRO leverages the reasoning capabilities of a VLM and encourages it to first identify objects relevant to the gaze-based task before inferring user intent. To extract objects of interest, \ACRO employs point-query-based segmentation using SAM2 \cite{ravi2024sam} to generate object masks corresponding to the user's gaze points.
Given gaze points specifying targets of interest, we extract corresponding depth frames and instance segmentation maps from each of the two robot-mounted cameras. To accommodate potential inaccuracies in gaze estimation, we select the nearest segmented object mask within a predefined margin around each gaze point. The resulting object point clouds from both cameras are then merged into a unified coordinate system aligned with the robot's base frame, providing an accurate and cohesive representation of the user's intended targets.

\noindent\textbf{Grasp Pose Generation.}
To facilitate gaze-driven pick-and-place manipulation, we implemented a robust grasp generation pipeline based on Contact-
GraspNet~\cite{sundermeyer2021contact} and a dual RGB-D camera setup.
Contact-GraspNet's predictions depend heavily on the viewpoint used during training, typically optimized for top-down perspectives. To address the issue of incorrect grasp orientations (e.g., grasps predicted beneath objects), we augmented our input data by rotating the merged point cloud about the X-axis at three discrete angles (180°, +135°, -135°), simulating natural top-down viewpoints. For each rotated point cloud, grasp candidates were confined to a bounding box around the segmented object and further filtered to ensure proper contact with the segmented region. From the filtered grasps, we select the median pose for robustness and further diversified the grasp set by applying ±45° pitch adjustments.
In cases where none of the rotations yielded a valid grasp candidate, the grasp closest in Euclidean distance to the original gaze point was selected from all predicted grasps. This comprehensive approach enhances viewpoint invariance, improves grasp orientation accuracy, and ensures reliable and diverse grasp predictions.

\begin{table}[]
    \centering
    \renewcommand{\arraystretch}{1.2} 
    \begin{tabular}{c|ccccc}
        \hline
        Task / VLM &  Gemini2.0F & Gemini2.5F & Gemini Pro & Llama4-Maverick & GPT-4o\\ \hline



       Lab-Tabletop & 0.93 & 0.91 & 0.94 & 0.75  & 0.78 \cr
         Wild (DROID) & 0.64 & 0.67 & 0.73 & 0.37 & 0.73 \cr
         \midrule
       Average &  0.79 & 0.79 & \textbf{0.84} & 0.56 & 0.76 \\ \hline
    \end{tabular}
    \vspace{0.1cm}
    \caption{VLM inference accuracies for predicting the user intent for 2 different datasets. For Lab-Tabletop tasks, each intent involves a sequence of actions and corresponding objects (e.g. pick up nail polish, place in basket). For Wild tasks, each intent is a natural language description.  }
    \label{tab:intent_inference}
    \vspace{-0.25cm}
\end{table}

\subsection{Reasoning with VLMs}
\ACRO is designed to enable zero-shot manipulation by using VLM reasoning at both the task level and grasp selection level. 
For high-level intent prediction, we design both the visual gaze prompt and the text prompt to encourage chain-of-thought reasoning within the model output. 
For the low-level contextualized grasp prediction, we design visual prompts to allow VLMs better understand grasp pose, and utilize more specialized reasoning models that does retrieval-augmented reasoning before generating a response.

\subsubsection{\textbf{Gaze-Based Intent Prediction.}}
With the predicted gaze point per timestep, we detect user gaze fixations with a maximum pixel distance threshold ($\delta=15p$) and a minimum temporal duration threshold ($t=2s$). 
Given the predicted gaze fixations on the image from the robot's camera view, we visualize the center of the gaze as a solid color dot and annotate the order of each gaze by directly placing text 1 or 2 next to the gaze point. 
We use the annotated image as the visual prompt and design a text prompt for intent prediction. Specifically, we instruct the VLM to first detect what objects are near the gaze points and then reason about what the user would want to do with these objects.
To evaluate intent reasoning with annotated gaze fixations, we constructed 2 datasets. We first designed 30 high-level intent recognition task scenarios for tabletop manipulation with different difficulty levels. 
The VLM is provided with possible robot primitive actions as options. We also sampled and annotated 45 task scenes from the DROID dataset~\cite{khazatsky2024droid}, which contains in-the-wild manipulation tasks. We randomly sampled a set of 10 language instructions to form a multiple choice question. Example tasks of both datasets are visualized in Figure \ref{fig:intent-reasoning}.
We evaluated 5 candidate VLMs with the same prompts in order to pick the best-performing model (more details in Section \ref{sec:experiments}).
Figure \ref{tab:intent_inference} presents the results of the evaluation. Subsequently, we pick the Gemini Pro model as the default intent prediction VLM to maintain highest performance.

\begin{figure}
    \centering
    \includegraphics[width=\linewidth]{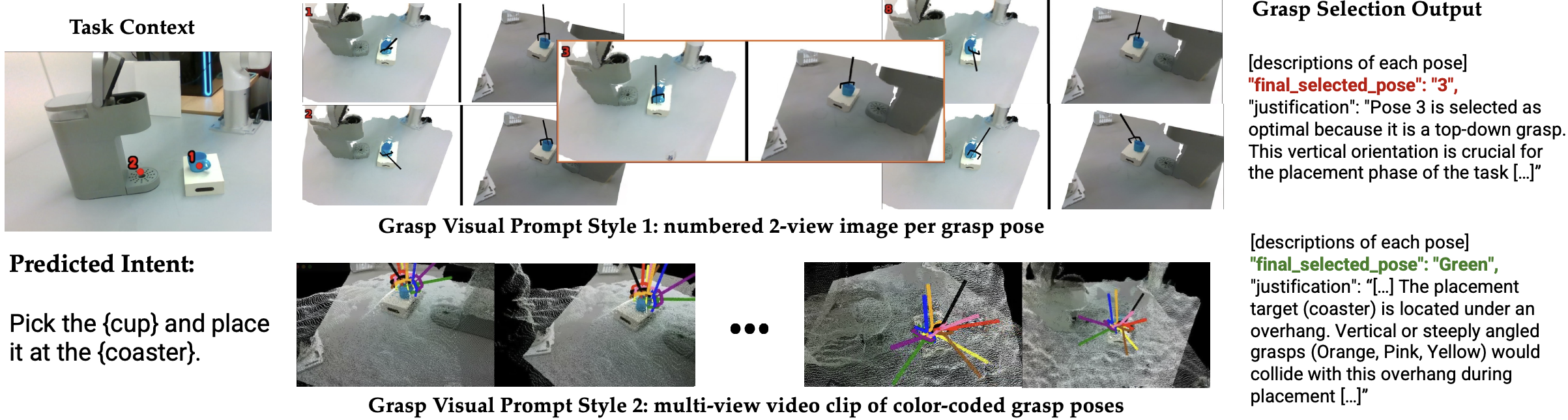}
    \caption{\textbf{Grasp Selection Visual Prompts.} We evaluated with different visual representations for grasp selection. To provide enough information for inferring the 3D grasping poses, we use both numbered multi-view image prompts (top) and a short video clip of a camera hovering around color-coded grasp pose candidates (bottom). Different visual representation resulted in the VLM (Gemini 2.5 Pro) making different predictions. }
    \label{fig:grasp-selection}
\end{figure}


\begin{table}[]
    \centering
    \renewcommand{\arraystretch}{1.2}
    \begin{tabular}{l|cccccc}
    \hline
    \textbf{Prompt / VLM} & Gemini2.0F & Gemini2.5F & Gemini Pro & Llama4-Maverick & GPT-4o \\
    \hline
    Image Average   & 0.20 & 0.47   & \textbf{0.60}     & 0.13 & 0 \\
    Image Time      & 6.42 & 5.44  & 24.83  & 5.41 & 8.18 \\ \hline
    Video Average   & 0.27 & \textbf{0.67}   & 0.47    & -- & -- \\
    Video Time      & 15.96 & 15.18 & 32.06  & -- & -- \\
    \hline
    \end{tabular}
    \vspace{0.1cm}
    \caption{VLM grasp selection performance across visual prompts. "--" indicates missing or unsupported data.}
    \label{tab:grasp_selection}
    \vspace{-0.25cm}
\end{table}

\noindent\textbf{Grasp Pose Selection.} Selecting an appropriate grasp pose is critical for successful object manipulation. Grasp pose prediction has long been a central research topic in robotics \cite{du2021vision, sundermeyer2021contact}. Vision-based grasp prediction models propose gripper poses that can securely grasp and lift objects based on geometric information from point clouds. However, task success often depends not only on stability of the grasp but also on contextual factors. For example, to place a cup into a microwave, one must grasp it from the side to avoid collisions with the microwave walls.
To incorporate environmental context into grasp pose selection, \ACRO leverages the reasoning capabilities of vision-language models to infer potential challenges, such as collision risks, associated with each candidate grasp. 

Specifically, \ACRO employs an existing vision-based grasp prediction model, Contact-GraspNet \cite{sundermeyer2021contact}, to generate a set of 9 distinct potential grasp poses at varying angles for a given object pointcloud.
To facilitate grasp evaluation, \ACRO implements two main types of visual prompts for the VLMs. 
For VLMs that accept image inputs, \ACRO compiles a prompt for each of the 9 candidate grasps with 2 different camera views, labeling each with a unique identifier (\cref{fig:grasp-selection} top). For VLMs that cannot accept all 9 images (Llama), we concatenate the individual images into a grid.
For VLMs capable of processing videos (i.e. Gemini models), \ACRO renders a short clip showing a 3D visualization of multiple grasp options overlaid on the scene's point cloud, viewed from different camera angles (\cref{fig:grasp-selection} bottom). The video clip consumes more tokens but provides a more comprehensive visualization of the grasping pose.

The grasp selection prompt first uses the output of the intent prediction to provide the high-level context of the manipulation task, then instructs the reasoning model to describe each candidate grasp in extensive detail, including potential issues such as stability, collision risk, or reachability. Based on this analysis, the model identifies all valid grasp options, filtering out any infeasible ones. Finally, considering both the task requirements and the desired final object placement, the model selects the single best grasp pose for execution.
We evaluated 5 candidate VLMs for grasp selection using different visual prompt styles and present the result in \cref{tab:grasp_selection}. Gemini Pro achieved the highest success rate picking the grasp pose with image prompts, with a cost of relatively long inference time. 
Based on this offline benchmarking, we identify Gemini 2.5 Flash with video prompts as the best-performing configuration for grasp selection. However, to ensure consistent system behavior during human evaluation, the user study experiments were conducted with a fixed grasp selection VLM (Gemini 2.5 Pro) chosen prior to this ablation. Consequently, the performance metrics reported in \cref{sec:experiments} represent a conservative evaluation of the system, as the optimizations found in the ablation study were not yet integrated into the real-time execution loop at the time of the trials.

\begin{figure}
    \centering
    \includegraphics[width=\linewidth]{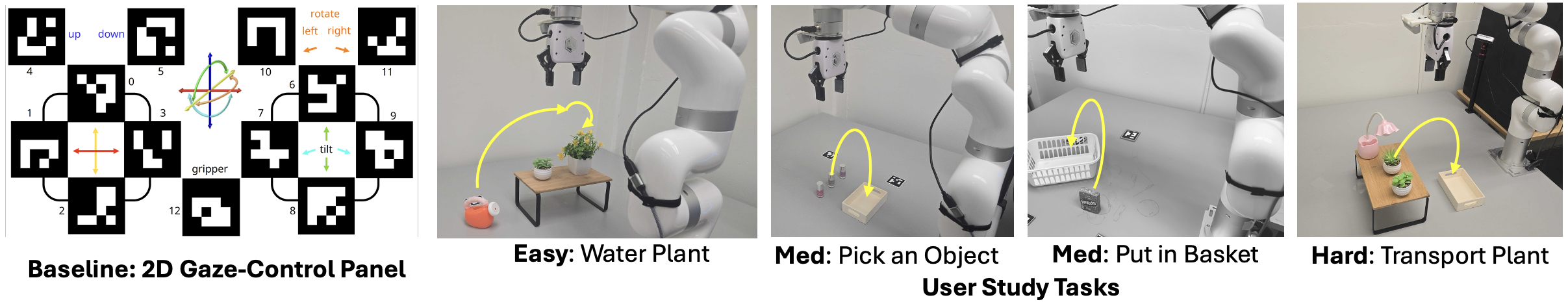}
    \caption{Experimental setup. Left: baseline 2D gaze control panel. Right: tasks in user study are of different difficulty levels. The more constrained the picking pose is, the harder the reasoning and execution becomes. }
    \label{fig:experiments}
    \vspace{-0.25cm}
\end{figure}

\section{Experiments}
\label{sec:experiments}


We evaluate \ACRO with a diverse set of table-top manipulation tasks, using a Ufactory Xarm 7 robotic arm (see~\cref{fig:experiments}). We use smart glasses from Meta's Project Aria~\cite{projectaria} for eye tracking and egocentric gaze prediction. The robot has access to 2 Realsense RGB-D cameras to observe the scene. We designed benchmarking tasks for each module, 
and conducted a user study to collect subjective feedback from non-expert users.

\noindent \textbf{VLM Reasoning Evaluations.}
To inform our design choices for \ACRO, we evaluate state-of-the-art VLMs for both high-level task planning and low-level grasp selection tasks. Specifically we consider 3 different versions of Gemini~\cite{team2023gemini} (2.0-Flash, 2.5-Flash, and Pro), Llama4 Maverick~\cite{meta_llama4}, and GPT4-o~\cite{achiam2023gpt}. We designed 30 different scenes as test cases for each of the reasoning task as previously described in \cref{sec:method}. Note that challenging test cases for the two tasks are different. Intent recognition tests VLMs' capability for object recognition and common sense reasoning, where clutter and adversarial visual attacks can be challenging. For grasp selection, the most challenging aspect is to have a good visual understanding of the grasping pose and analyze the scene geometry to take both task semantics and obstacle avoidance into consideration. 
We observe that Gemini models perform better than alternatives. 

\noindent \textbf{User Study.}
To evaluate the effectiveness of \ACRO in assisting users that are unfamiliar with the system's internal workings, we conducted a user study comparing \ACRO with a baseline panel-based gaze-control method (see \cref{fig:experiments} left). While there is no existing assistive manipulation system implement the exact approach, design of this baseline is inspired by gaze-based typing systems~\cite{ward2000dasher,feng2021hgaze}. 
With the gaze-panel method, users control the 6-DoF pose of the robotic arm and the gripper by selecting virtual buttons (visual markers) on a screen using their eye gaze. 
This comparison allows us to assess the intuitiveness, efficiency, and usability of \ACRO's VLM-powered intent inference versus more conventional gaze-based interfaces.

\begin{itemize}
[itemsep=0.2em,nolistsep,labelindent=0.5em,labelsep=0.15cm,leftmargin=*]
\item \textbf{Participants}: We recruited 6 college students (4 male, 2 female) to participate in our study, with an age range 20-28. Two participants indicated having prior experience interacting with robots.

\item \textbf{Independent variables}: Each participant interacted with two modes of gaze-based control---baseline gaze-panel controller (Gaze Panel), and our system \ACRO.

\item \textbf{Dependent variables}: We take objective measures by logging the success rate of each method at various stage of a task, as well as the time spent by the user to complete a task; we also collect subjective feedback from the users by asking them to fill out a likert-scale survey adapted from the NASA TLX questionnaire~\cite{hart2006nasa}. 

\item \textbf{Hypotheses}: We hypothesize that \textbf{H1}) \ACRO achieves similar performance as the baseline in terms of success rate while requiring less time and effort than baseline, and \textbf{H2}) \ACRO is preferred by users for its ease of use.

\item \textbf{Procedure}: Each participant was instructed to first use their natural gaze to communicate intent while their eye gaze was recorded. They were then introduced to our gaze control methods and given three minutes to practice each. During the evaluation, participants completed four tasks using both control methods, with up to two trials per task. Task examples are shown in \cref{fig:experiments}. Notably, the \textit{Put in Basket} task required grasp poses that avoided the basket walls, and the \textit{Transport Plant} task required avoiding a lamp positioned above the plants. After each task, participants completed a Likert-scale survey reflecting on their mental and physical effort.

\end{itemize}

\begin{figure}
    \centering
    \includegraphics[width=\linewidth]{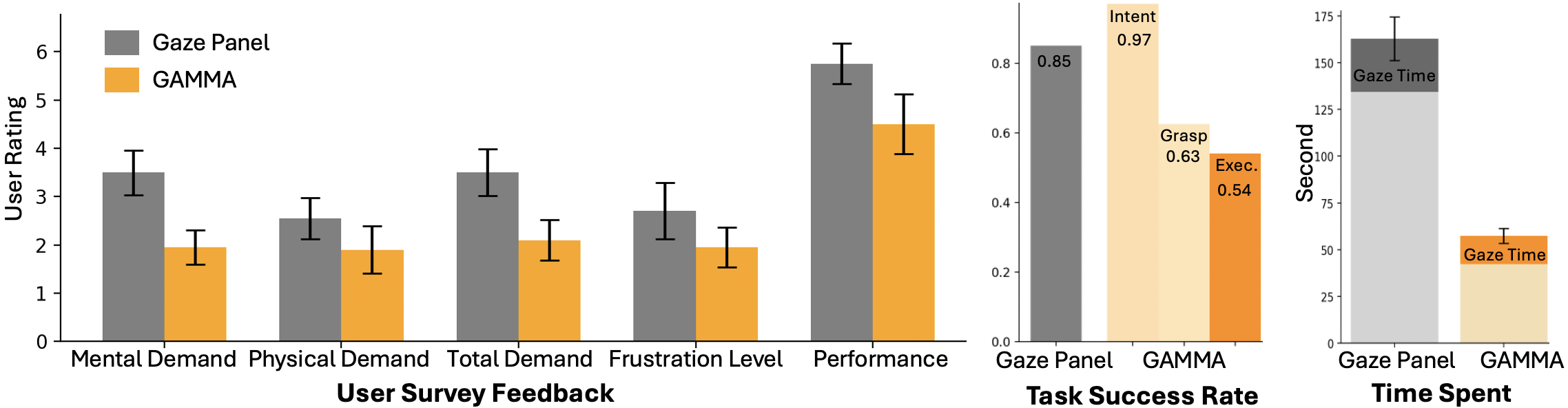}
    \caption{\textbf{User Study Results.} We present the subjective evaluation in average likert-scale ratings from the users and the objective measure of time spent and success rates. We see that while users consider \ACRO to require lower demand and spent less time on the task, their performance is higher when they have the full control of the robot as \ACRO often requires another trial to correct a wrong prediction. }
    \label{fig:user-results}
    \vspace{-0.25cm}
\end{figure}

\textbf{Results.} 
We first analyzed users' natural gaze patterns and found that \ACRO's fixation-based model can effectively interpret intent for 5 out of 6 participants. One participant exhibited additional referential gaze behaviors directed toward the robot—likely to monitor or coordinate its actions. This analysis supports the generalizability of our gaze-based intent inference approach while highlighting the importance of accommodating diverse gaze patterns across users.

\begin{wrapfigure}{r}{0.39\textwidth}
  \centering
  \includegraphics[width=0.39\textwidth]{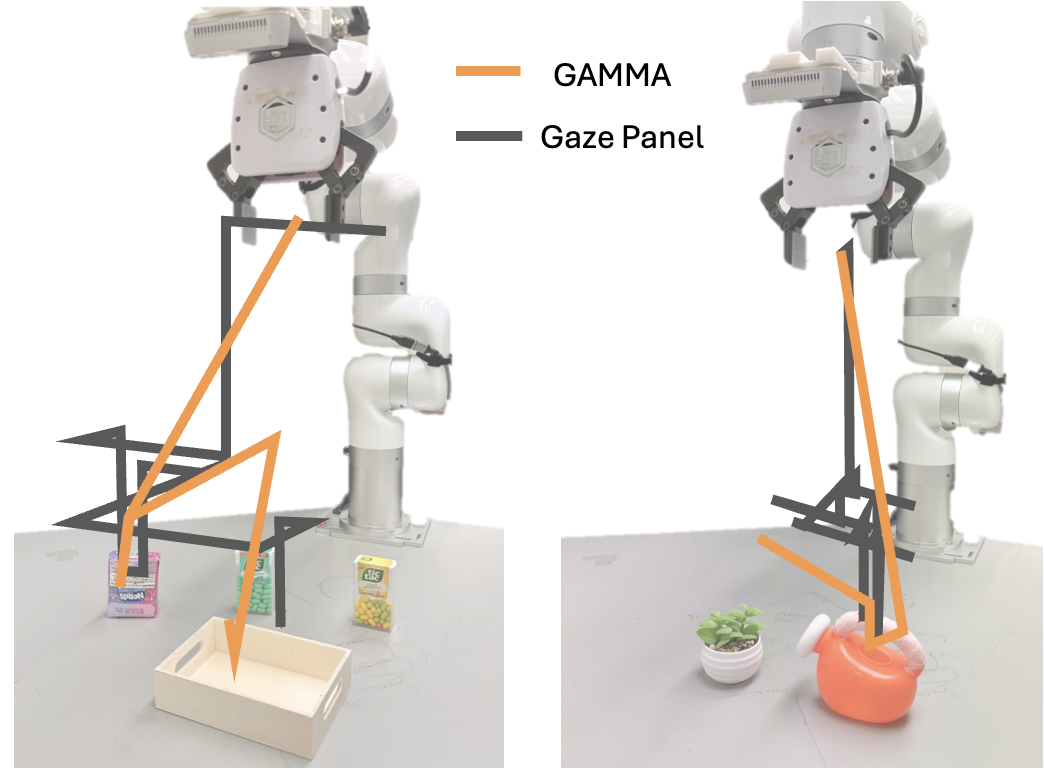}
  \caption{\textbf{Visualization of sample trajectories.} \ACRO trajectories are shorter and cleaner than gaze panel trajectories.}
  \label{fig:visualization}
\end{wrapfigure}
We present both objective performance metrics and subjective user feedback in \cref{fig:user-results}. Overall, \ACRO required less cognitive effort and elicited lower frustration compared to the gaze-panel baseline. Interestingly, users rated the gaze panel as having higher perceived performance, largely because its direct control allowed them to actively recover from mistakes. While \ACRO achieved a high success rate in inferring user intent, it did not consistently produce reliable grasps, which is a limitation stemming from compounding errors in both the zero-shot grasp predictor and the VLM-based grasp selection module. Notably, four out of six participants ultimately preferred the gaze-panel interface, citing \textit{“having more direct control,”} \textit{“feeling more interactive,”} and being \textit{“more consistent”} as factors.

As shown in the rightmost plot of \cref{fig:user-results}, \ACRO enabled users to complete tasks significantly faster than the gaze panel (p-value $\ll 0.01$). Sample user-study trajectories are illustrated in \cref{fig:visualization}. Interestingly, in the gaze-panel condition, participants spent more time thinking and mentally planning than physically controlling the robot, as indicated by the darker gray regions representing gaze time.

Our findings support \textbf{H1} (performance and efficiency) but do not support \textbf{H2} (user preference). We attribute the rejection of \textbf{H2} to two interacting factors: (1) the execution risk and failure rate associated with zero-shot motion generation, and (2) a nuanced trade-off between ease of use and perceived agency. One-shot success in zero-shot LLM-as-policy systems remains unreliable and often demands situated learning or post-deployment human guidance to ensure successful execution~\cite{zha2024distilling}. When unexpected failure occurs, it is not easily recoverable with a fully autonomous pipeline. 
Simultaneously, prior work such as that of \citet{bhattacharjee2020more} has shown that users—particularly those without substantial mobility impairments—frequently prefer systems that preserve a greater sense of agency and control. Together, these observations highlight the need for assistive interfaces that maintain an appropriate balance between autonomy and meaningful user involvement, aligning system behavior with user expectations and preferences.

\noindent\textbf{Summary.} In this work, we present \ACRO, a framework for interpreting human gaze as manipulation commands for assistive robots. \ACRO leverages state-of-the-art vision and vision-language foundation models to enable zero-shot task execution. We demonstrate the effectiveness of \ACRO through a series of table-top manipulation tasks, including watering plants, selecting objects from clutter, and making coffee.
We benchmark the performance of SoTA VLMs for gaze-guided task reasoning and conduct a user study to compare \ACRO with a panel-based gaze control baseline. Results show that \ACRO offers a more intuitive and lower-effort interaction experience. However, a majority of participants preferred the baseline method, citing a greater sense of agency and control over the robot's motion. These findings suggest that while automation is valuable, users also seek the ability to intervene and guide the robot's behavior more directly. A promising future direction is to design hybrid control mechanisms that allow users to fluidly switch between automated and manual control modes depending on task complexity and personal preference.

\noindent\textbf{Limitations.} Despite the promise of foundation models for enabling zero-shot robot manipulation, achieving smooth, responsive, and context-aware control over complex tasks remains a significant challenge. While our results show that VLMs are effective for high-level intent recognition and gaze interpretation, they continue to struggle with fine-grained, low-level reasoning essential for motion planning and physical interaction.
Furthermore, the computational cost of VLMs can lead to increased inference times, which negatively impacts system responsiveness and may contribute to user frustration. Improving the speed and precision of model reasoning remains a critical hurdle for designing fluid and effective human-robot interfaces.
Enabling mobile manipulation is a non-trivial generalization of \ACRO, which would require more advanced visual co-localization techniques beyond fixed ArUco markers, relying instead on natural scene features for camera pose estimation.
Finally, our user study was conducted exclusively with healthy participants. To more comprehensively evaluate the effectiveness and accessibility of systems like \ACRO, it is essential to include participants with limited mobility or physical impairments in future studies. This would provide deeper insights into the system’s usability and its potential impact in real-world assistive scenarios.

\noindent\textbf{Acknowledgment.}
The work is supported by Meta's Project Aria. This work is also partially funded by Google's Academic Research Gift.



\bibliography{references}  

\newpage
\appendix
\Large{\textbf{Appendix}}
\normalsize

\tableofcontents  
\vspace{0.25em}
\noindent\rule{\textwidth}{0.4pt}  
\vspace{0.5em}

\setcounter{figure}{0}
\renewcommand{\thefigure}{\arabic{figure}}

\setcounter{table}{0}
\renewcommand{\thetable}{\arabic{table}}

\section{Gaze Transformation Algorithm}

To transform the detected gaze point in Aria glasses (dynamic RGB camera) to the robot's (fixed RGB-D) camera, we project the 3D points from the RGB-D camera into the RGB camera and find the nearest neighbor to the 2D gaze point. The full algorithm is present in Algorithm 1.

\begin{algorithm}[H]
\caption{Gaze Transformation from RGB (Aria) to RGB-D (Realsense) Camera}
\KwIn{
    $D$ -- Depth image from RGBD camera,\\
    $K_{D}$ -- Intrinsics of RGBD camera,\\
    $T_{D \rightarrow R}$ -- Transformation from RGBD to RGB camera,\\
    $K_{R}$ -- Intrinsics of RGB camera,\\
    $p_{\text{gaze}}$ -- 2D gaze point in RGB image
}
\KwOut{$P_{3D}$ -- 3D point in RGBD camera space corresponding to gaze point}

\Begin{
    Initialize empty list $C_{\text{RGB}}$ for candidate 2D projections in RGB image\\
    Initialize empty list $C_{3D}$ for corresponding 3D points in RGBD frame\\

    \ForEach{pixel $(u,v)$ in $D$ with step size (e.g. 4)}{
        $Z \gets D[u,v] \cdot 0.001$ \tcp*{Convert depth to meters}
        \If{$Z == 0$}{
            \textbf{continue} \tcp*{Skip invalid or missing depth}
        }

        \tcp{Back-project depth pixel to 3D point in RGBD camera space}
        $X \gets \frac{(u - c_{x_D}) \cdot Z}{f_{x_D}}$ ~\hspace{0.2cm}
        $Y \gets \frac{(v - c_{y_D}) \cdot Z}{f_{y_D}}$\\
        $P_D \gets [X, Y, Z, 1]^T$ \tcp*{Homogeneous coordinates}

        \tcp{Transform point from RGBD to RGB camera frame}
        $P_R \gets T_{D \rightarrow R} \cdot P_D$ \hspace{0.2cm}
        $[X_R, Y_R, Z_R] \gets P_R[0:3]$

        \tcp{Project transformed 3D point to 2D in RGB image}
        $u_R \gets \frac{X_R \cdot f_{x_R}}{Z_R} + c_{x_R}$ ~\hspace{0.2cm}
        $v_R \gets \frac{Y_R \cdot f_{y_R}}{Z_R} + c_{y_R}$

        \tcp{Store the projected 2D point and original 3D point}
        Append $(u_R, v_R)$ to $C_{\text{RGB}}$\\
        Append $[X, Y, Z]$ to $C_{3D}$ \tcp*{Keep 3D in RGBD frame}
    }

    \tcp{Find the 2D projection closest to the gaze point}
    Find index $i$ of nearest neighbor in $C_{\text{RGB}}$ to $p_{\text{gaze}}$\\

    \tcp{Retrieve the corresponding 3D point in RGBD camera frame}
    $P_{3D} \gets C_{3D}[i]$\\
    \Return $P_{3D}$
}
\end{algorithm}

\section{Baseline Gaze-Panel Control}
We implemented a panel-based gaze control interface inspired by existing gaze-typing systems as a comparative baseline. This interface enables users to control the 6-degree-of-freedom pose of the robotic arm, as well as the gripper state, by fixating on designated virtual buttons rendered on a display screen. Each button corresponds to a discrete command, including translational and rotational adjustments along the x, y, and z axes, as well as open and close gripper actions. Eye gaze is captured in real-time using Aria glasses, and commands are executed continuously while the user’s gaze is maintained on a given button. If gaze shifts away from the button, motion immediately stops. This mechanism ensures precise user-driven control, allowing subjects to execute fine adjustments and correct errors directly. The robotic arm executes each selected command at a positional velocity of $60$ mm/s and a rotational speed of $15$ degrees/s, which are tuned to provide motion that feels natural and intuitive to subjects.

\section{Intent Recognition Tasks}
To evaluate the VLM’s capacity for gaze-based intent recognition, we designed 30 task scenarios across 3 tasks, structured to probe semantic understanding and reasoning over gaze-driven inputs. Task difficulty was modulated by varying the number of objects, introducing scene clutter, adjusting viewpoint complexity, and embedding visually similar distractors. Each scenario was built to challenge the VLM’s ability to infer likely user goals given two or more gaze fixations within a visually complex environment.

\begin{figure}[h!]
    \centering
    \includegraphics[width=\linewidth]{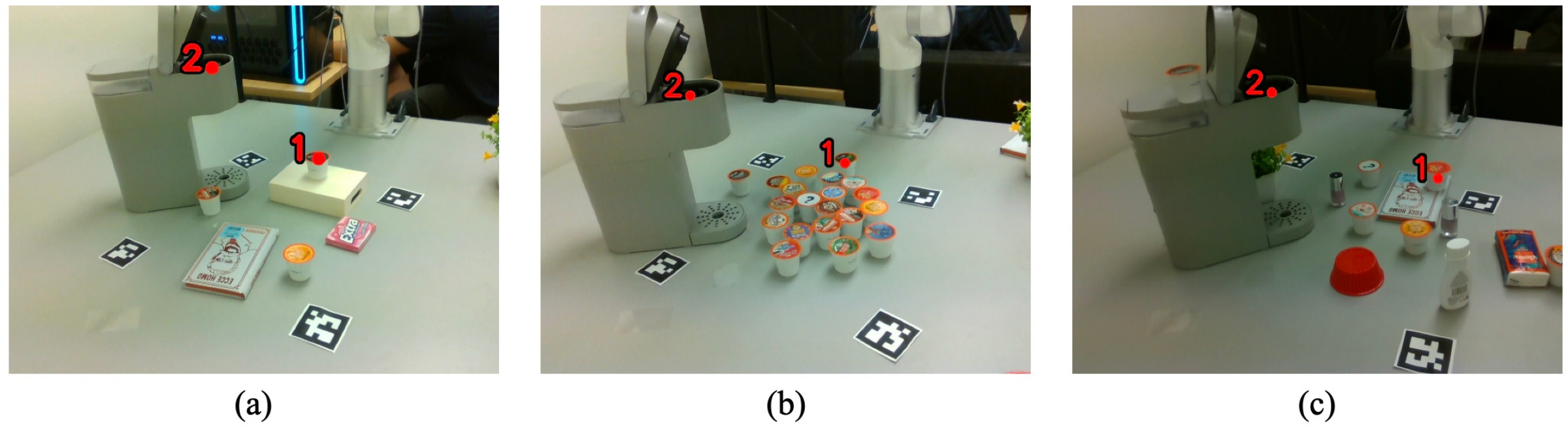}
    \caption{\textbf{Example scenarios used in the intent recognition task.} (a) An example of easy tasks. (b) An example of medium-difficulty tasks. (c) An example of hard tasks.}
    \label{fig:intent-examples}
    \vspace{-0.25cm}
\end{figure}

For instance, as shown in Figure \ref{fig:intent-examples}, an easy task might involve gazing first at a coffee pod and then at a coffee maker, implying a direct, familiar operation like “pick the pod and place into the machine.” Medium-difficulty tasks often required the model to reason through object identity and exclude confounding visual cues. In one such example, the coffee pod was placed among several visually similar items, requiring the model to correctly identify the intended object despite high local similarity and spatial crowding. Hard tasks introduced further ambiguity, including partial occlusion or low visual salience. For instance, a coffee pod might be partially obscured by other objects or placed on a surface with nearly identical coloration, such that only shape or minimal texture contrast could guide recognition. These hard examples challenged the model not only to detect target objects under perceptual constraints but also to maintain goal-consistent reasoning when gaze sequences were ambiguous or misleading.

To support structured reasoning in such tasks, we designed a custom prompt for the VLM that decomposes intent inference into discrete steps. This prompt first instructs the model to localize the gaze targets using a constrained object vocabulary and then infer the most plausible action between them using a limited action set (e.g., pick\_and\_place, pouring). The VLM input consists of an annotated image from a robot camera, where gaze fixations are shown as numbered red dots overlaid on the visual scene. This image, paired with the structured prompt, serves as a visual-language query, guiding the model to first identify the objects under each gaze point and then determine a likely interaction based on context.
\section{Pose Selection Tasks}
To systematically evaluate our grasp pose selection system, we designed three representative pick-and-place scenarios that emphasized distinct environmental and interaction constraints. The objective was to select a single optimal grasp pose that would allow the robot to execute both picking and placing without requiring any reorientation of the end effector after grasping. This constraint underscores the importance of selecting a grasp that is not only stable and reachable at pickup but also suitable for the final placement configuration.

\noindent \textbf{Multi-object scene}
This scenario involved several objects arranged in close proximity. It evaluated the model's ability to correctly interpret gaze intention and disambiguate among candidate objects. It also tested the model's reasoning about nearby distractors that could interfere with grasp execution or placement.

\noindent \textbf{Basket placement scene}
Here, the object had to be placed inside a basket. Some objects were positioned near the basket walls, making top-down grasps unsuitable due to potential collision during placement. The task required the model to identify viable side-approach grasps when placing, without excessive horizontal tilt that might cause collision.

\noindent \textbf{Overhead constraint scene}
In this task, an overhead obstruction limited the available vertical workspace near the placement area. The model needed to avoid top-down grasps, selecting a side approach compatible with the robot's reachability and placement constraints. Some variations of this scenario included multiple objects to increase task complexity.

For each scene, we provided the model with a high-level task description (e.g., “Pick the bottle and place it in the basket”). The model was instructed to analyze the options based on scene geometry, object shape, grasp stability, and kinematic feasibility. Prompts requested structured JSON outputs containing pose-by-pose reasoning, filtered valid grasps, and a final single pose selection. While multiple poses could be technically valid, we enforced selection of only one to assess the model's ability to prioritize under constraints.

To ensure reasoning quality, we included the camera view in the prompt to compensate for potential ambiguity due to the low-resolution point cloud visualization. Although higher-fidelity rendering was possible, we found it inefficient for the performance gain observed. Instead, we focused on clearly articulated prompt examples that emphasized core reasoning criteria. Each of the three tasks was instantiated in five different physical arrangements, producing a total of 15 scene variations as depicted in Table \ref{tab:scenarios}.

\footnotesize
\begin{table}[h!]
    \centering
    \renewcommand{\arraystretch}{1.2} 
    \begin{tabular}{l c c c c c}
        \textbf{Scene} & \textbf{Variation 1} & \textbf{Variation 2} & \textbf{Variation 3} & \textbf{Variation 4} & \textbf{Variation 5} \\
        \hline
        \shortstack[l]{\textbf{Multi-} \\ \textbf{object}}  &
        \includegraphics[width=0.15\textwidth]{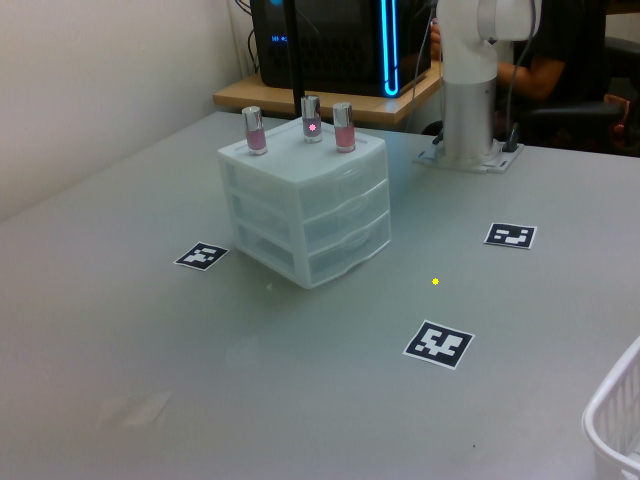} &
        \includegraphics[width=0.15\textwidth]{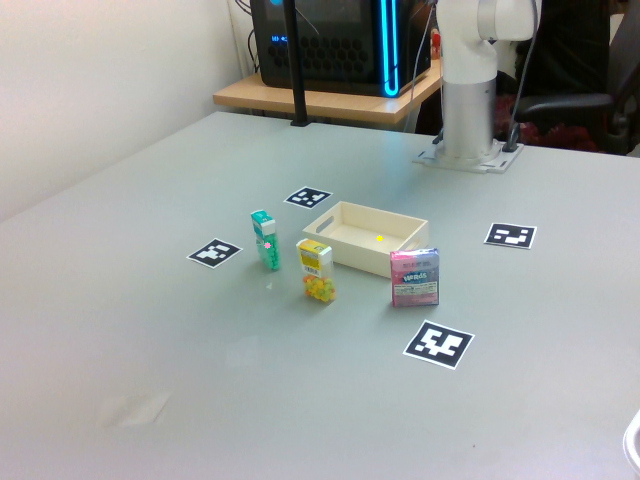} &
        \includegraphics[width=0.15\textwidth]{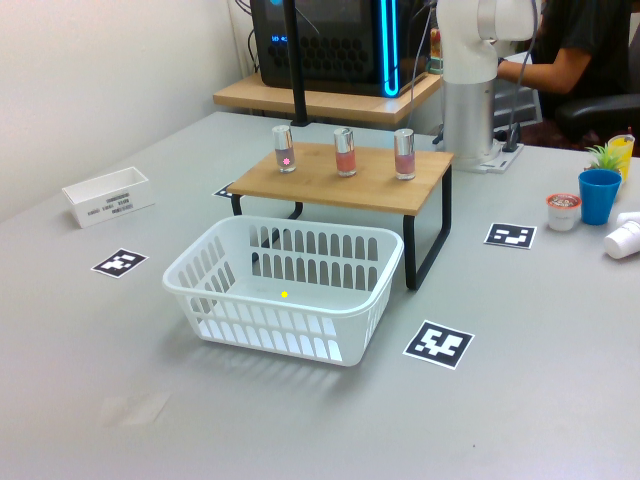} &
        \includegraphics[width=0.15\textwidth]{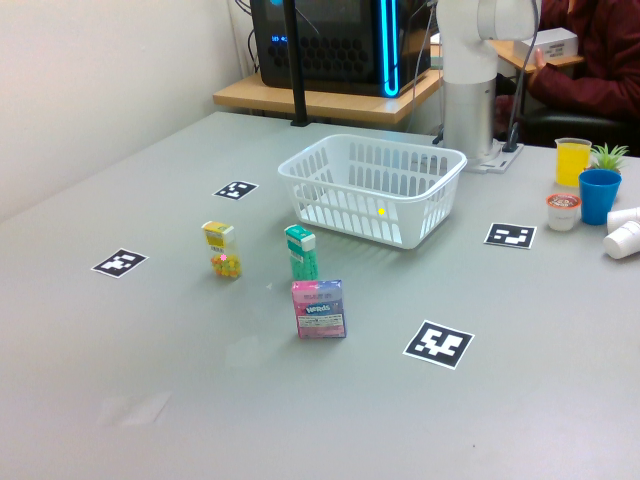} &
        \includegraphics[width=0.15\textwidth]{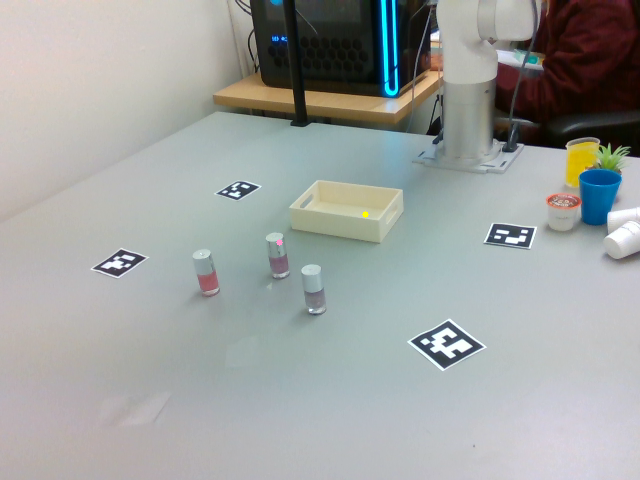} \\
        \hline
        \shortstack[l]{\textbf{Basket} \\ \textbf{placement}}  &
        \includegraphics[width=0.15\textwidth]{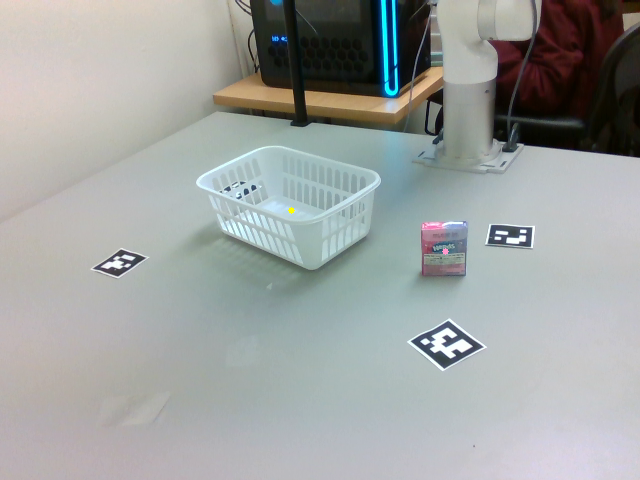} &
        \includegraphics[width=0.15\textwidth]{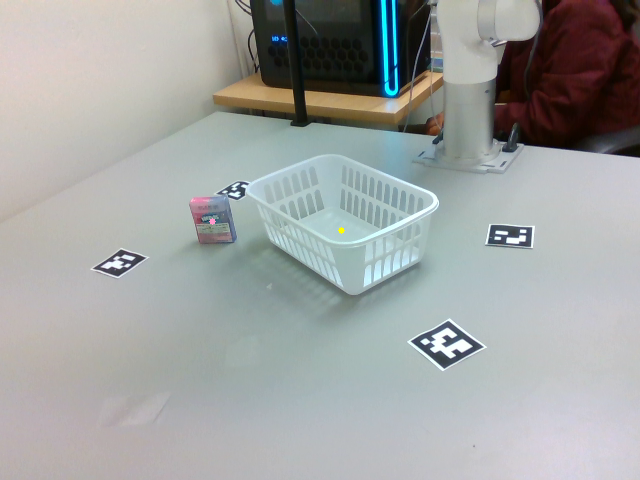} &
        \includegraphics[width=0.15\textwidth]{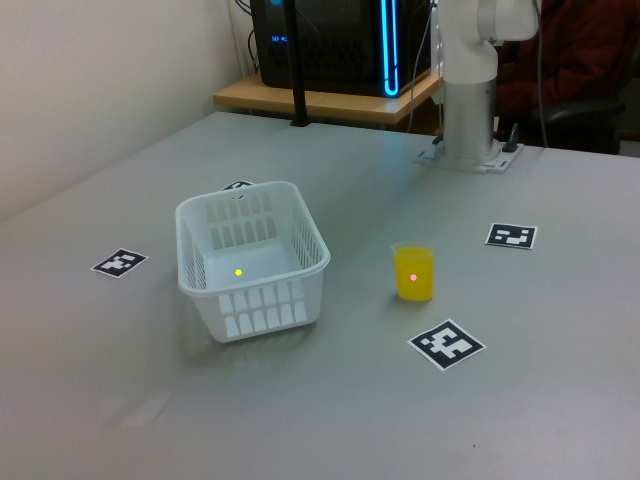} &
        \includegraphics[width=0.15\textwidth]{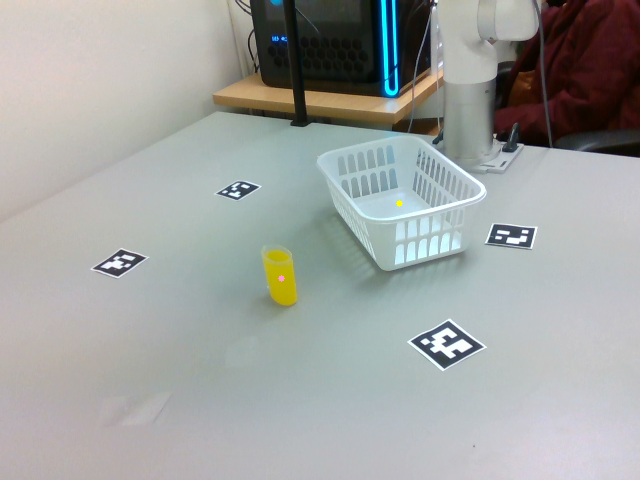} &
        \includegraphics[width=0.15\textwidth]{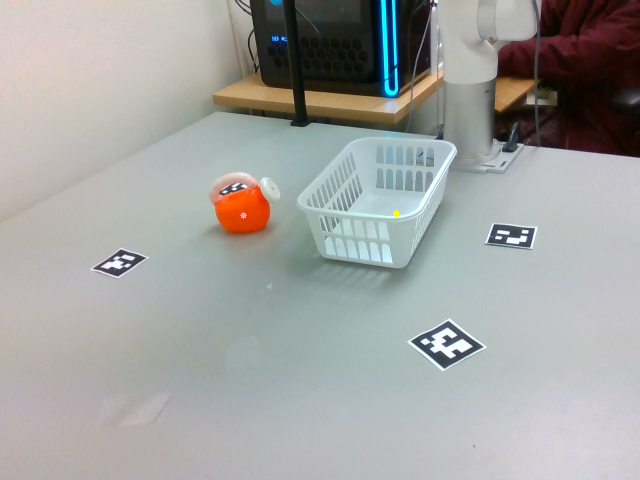} \\
        \hline
        \shortstack[l]{\textbf{Overhead} \\ \textbf{constraint}} &
        \includegraphics[width=0.15\textwidth]{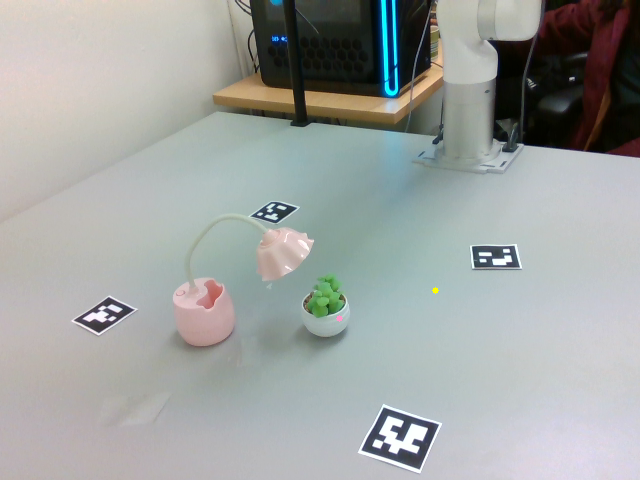} &
        \includegraphics[width=0.15\textwidth]{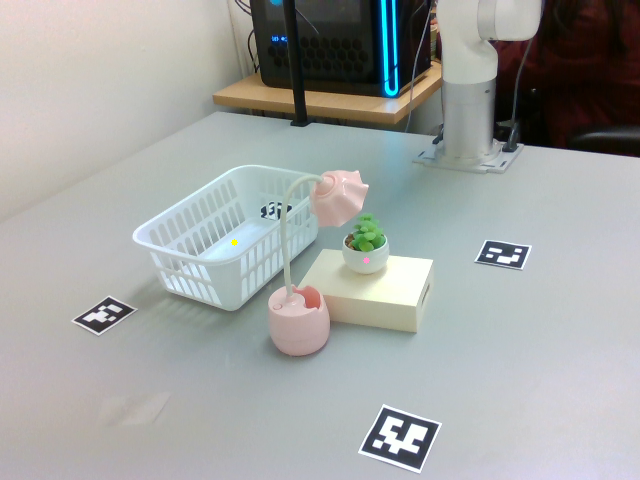} &
        \includegraphics[width=0.15\textwidth]{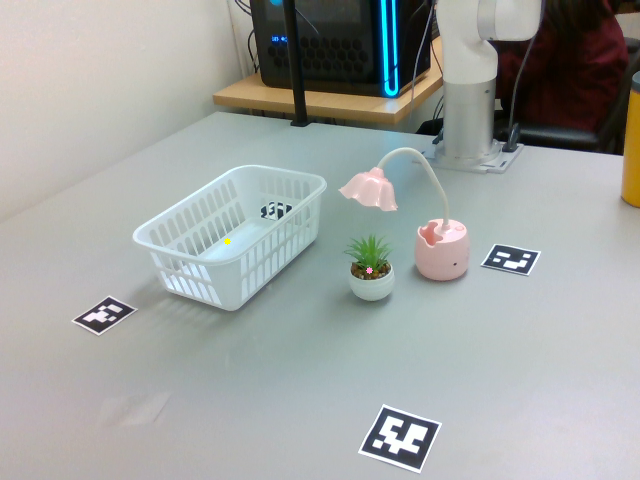} &
        \includegraphics[width=0.15\textwidth]{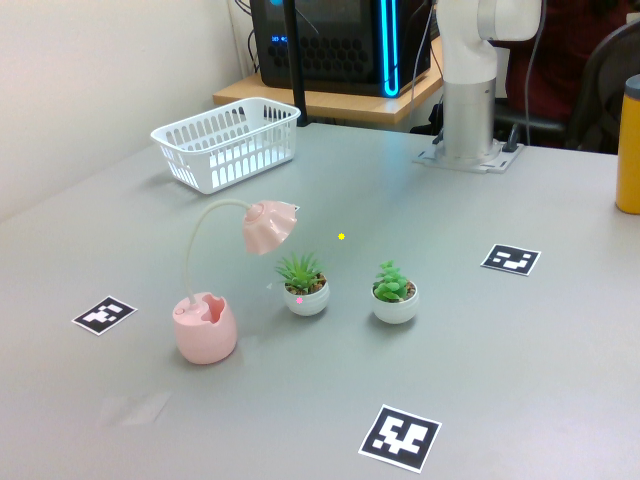} &
        \includegraphics[width=0.15\textwidth]{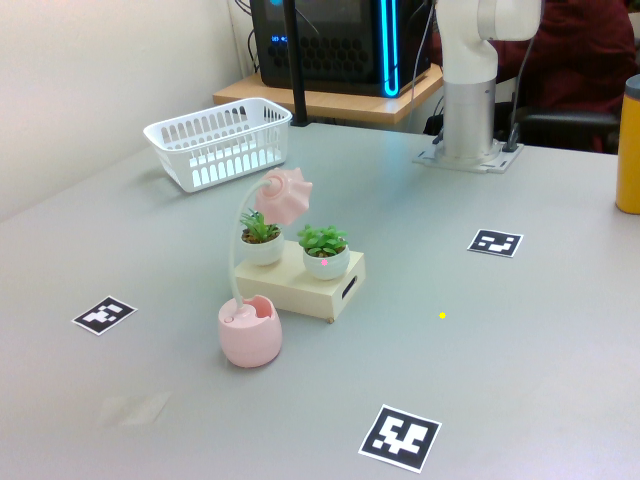} \\
        \hline
    \end{tabular}
    \caption{Grasp Pose Selection Scenarios}
    \label{tab:scenarios}
\end{table}

\begin{figure}[h!]
    \centering
    \begin{tabular}{|c|c|c|}
        \hline
        \includegraphics[width=0.3\textwidth]{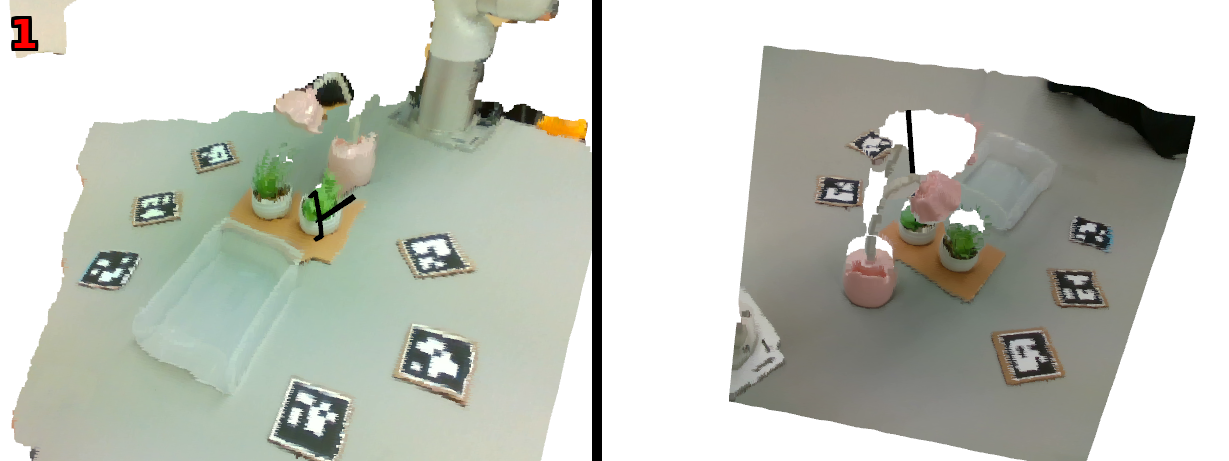} &
        \includegraphics[width=0.3\textwidth]{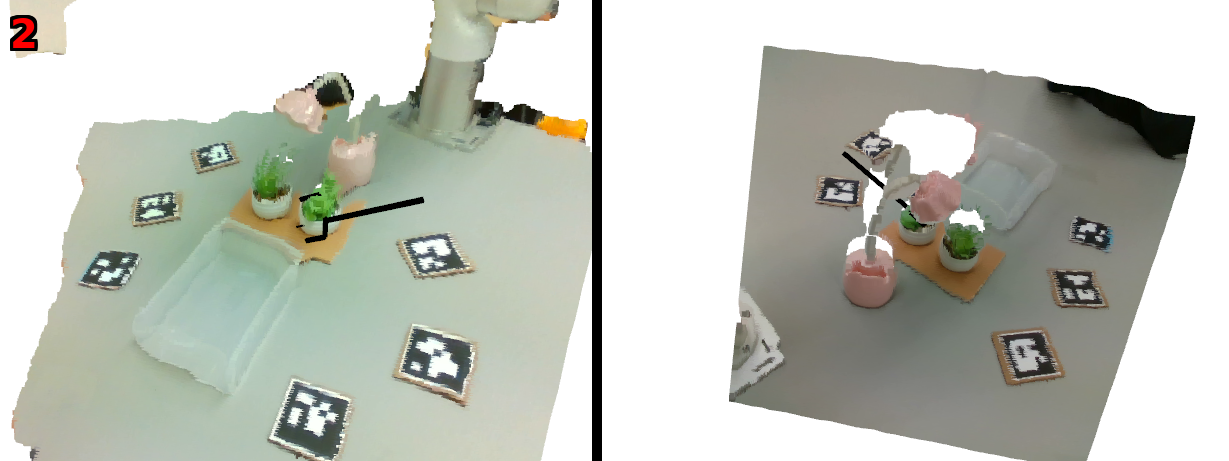} &
        \includegraphics[width=0.3\textwidth]{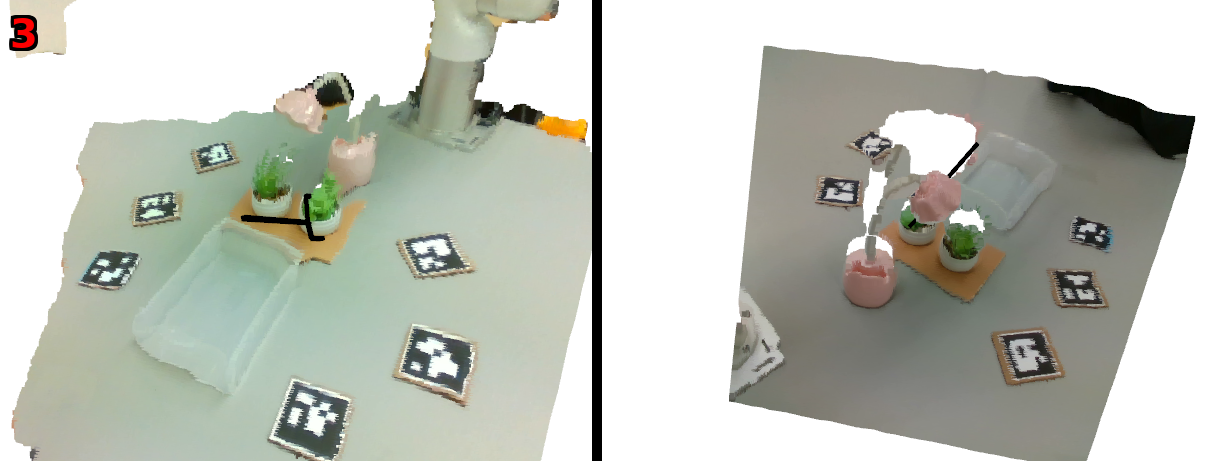} \\
        (a) Pose 1 & (b) Pose 2 & (c) Pose 3 \\
        \hline
        \includegraphics[width=0.3\textwidth]{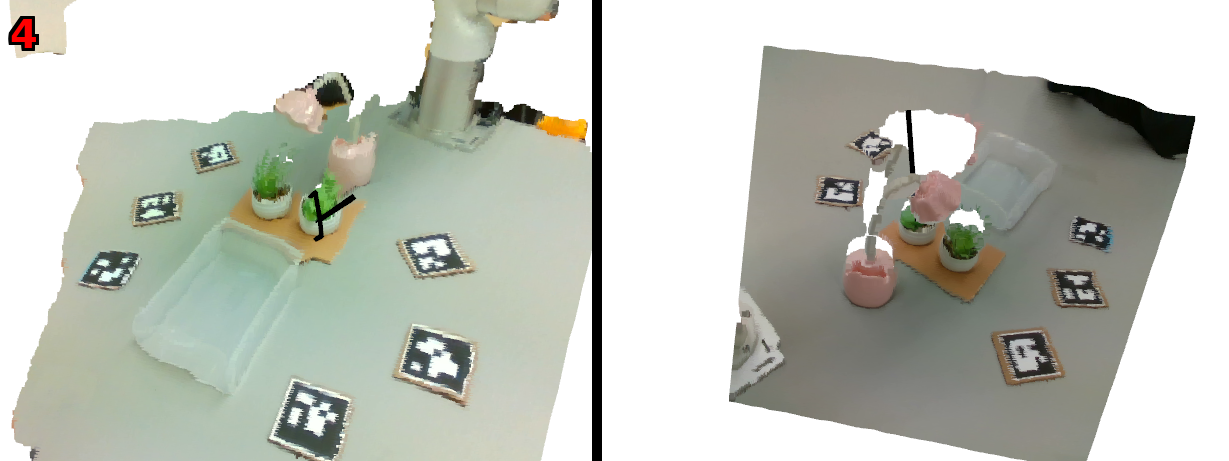} &
        \includegraphics[width=0.3\textwidth]{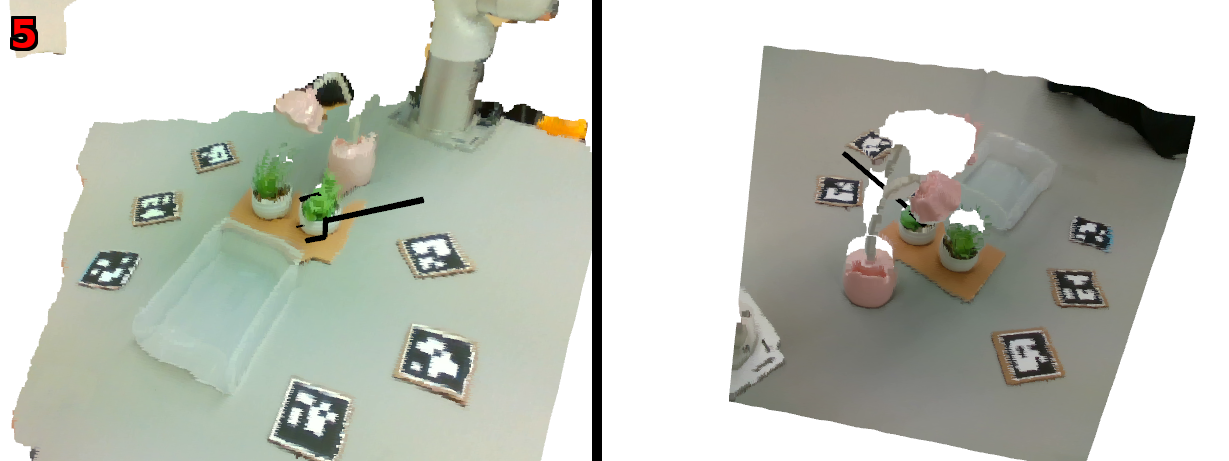} &
        \includegraphics[width=0.3\textwidth]{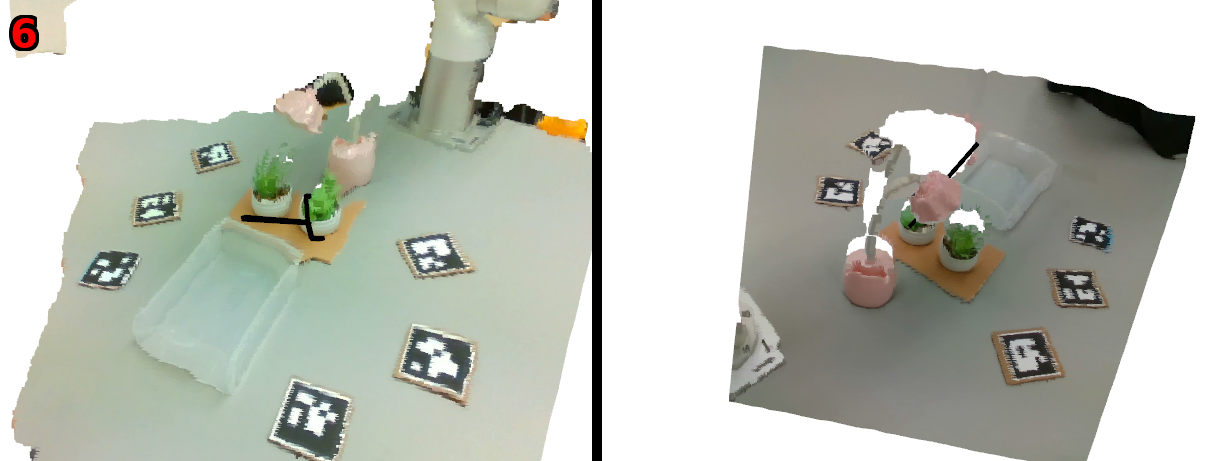} \\
        (d) Pose 4 & (e) Pose 5 & (f) Pose 6 \\
        \hline
        \includegraphics[width=0.3\textwidth]{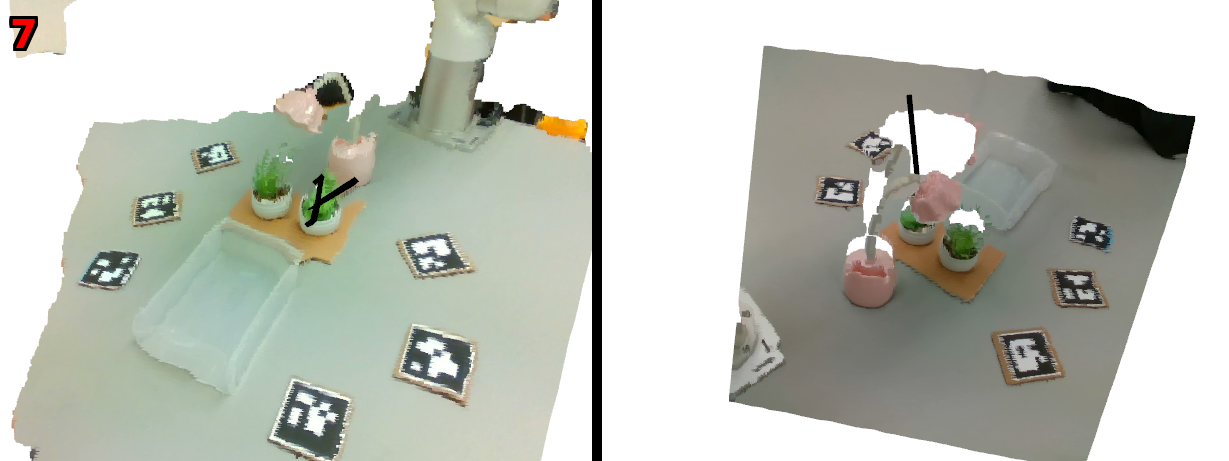} &
        \includegraphics[width=0.3\textwidth]{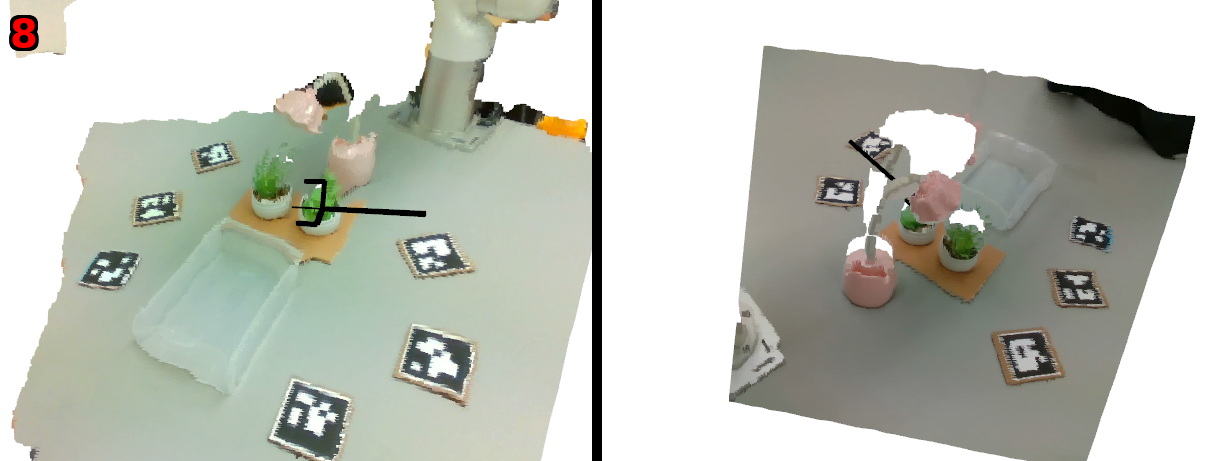} &
        \includegraphics[width=0.3\textwidth]{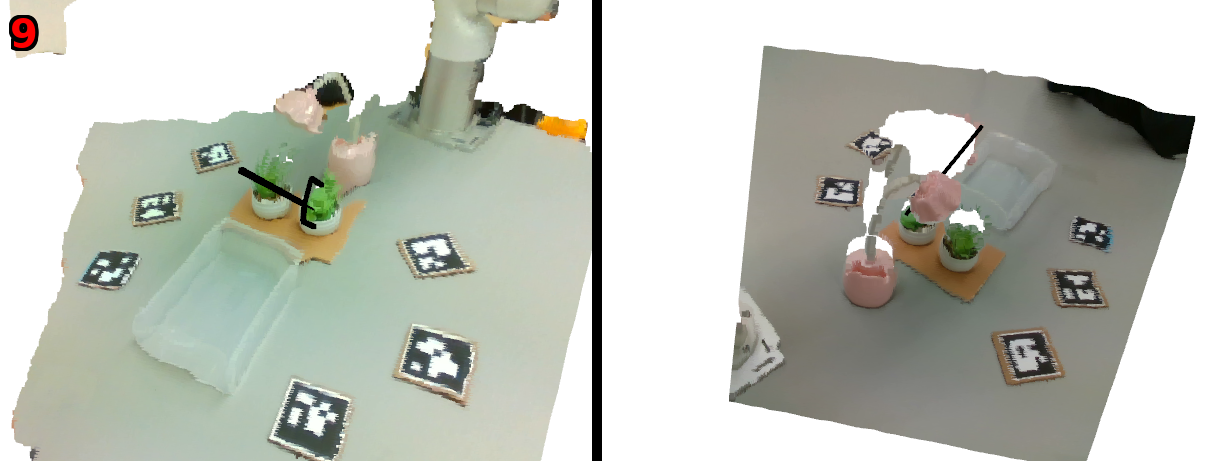} \\
        (g) Pose 7 & (h) Pose 8 & (i) Pose 9 \\
        \hline
    \end{tabular}
    \caption{Candidate Grasp Poses}
    \label{fig:grasp_poses}
    \vspace{-0.25cm}
\end{figure}

\begin{figure}[h!]
    \centering
    \includegraphics[width=0.5\textwidth]{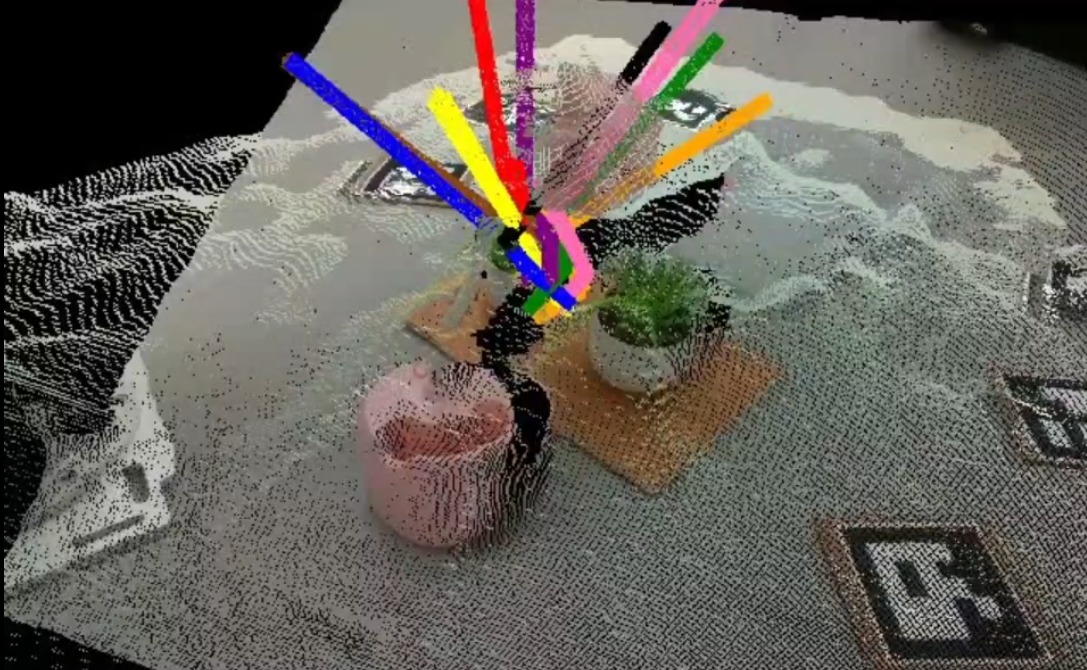}
    \caption{Screenshot of Pose Selection Video}
    \label{fig:pose_video}
    \vspace{-0.25cm}
\end{figure}

\normalsize
Two visualization methods were used as prompts for the VLM in our comparison. The first method presented images of each of the nine grasp poses, viewed from two different camera angles, as illustrated in Figure~\ref{fig:grasp_poses}. This approach required the VLM to process nine pairs of input images. The second method involved feeding a video of the grasps, where individual poses were distinguished by different colors, with these colors mapped to the respective poses shown in Figure~\ref{fig:pose_video}.

\section{User Study Questionnaire : Task \& Overall Evaluation}

\subsection*{Task Experience Evaluation (Per Task \& Mode)}

\textit{Scale: 1 (lowest) – 7 (highest)}

\begin{itemize}[leftmargin=2em]
    \item For each task (Pour, Multiple Objects, Place into Basket, Overhead Element) and each control mode (Panel, Pointer), please rate the following:
    \begin{itemize}
        \item Mental Demand: (1-7)
        \item Physical Demand: (1-7)
        \item Total Demand: (1-7)
        \item Performance: (1-7)
        \item Frustration: (1-7)
    \end{itemize}
\end{itemize}

\subsection*{Overall Evaluation}

\begin{itemize}[leftmargin=2em]
    \item \textbf{Overall, which method do you prefer?}
    \begin{itemize}
        \item A. Panel
        \item B. Pointer
    \end{itemize}

    \item \textbf{Could you please share the reasons for your preference?}
    \vspace{1em} \hrulefill \vspace{1em}

    \vspace{-0.45cm}
    \item \textbf{If you were disabled and could only move your eyes, which method would you prefer?}
    \begin{itemize}
        \item A. Panel
        \item B. Pointer
    \end{itemize}
\end{itemize}

\end{document}